\documentclass[pdflatex,sn-mathphys-num]{sn-jnl}% Math and Physical Sciences Numbered Reference Style 
\usepackage{graphicx}%
\usepackage{multirow}%
\usepackage{amsmath,amssymb,amsfonts}%
\usepackage{amsthm}%
\usepackage{mathrsfs}%
\usepackage[title]{appendix}%
\usepackage{xcolor}%
\usepackage{textcomp}%
\usepackage{manyfoot}%
\usepackage{booktabs}%
\usepackage{algorithm}%
\usepackage{algorithmicx}%
\usepackage{algpseudocode}%
\usepackage{listings}%
\theoremstyle{thmstyleone}%
\theoremstyle{thmstyletwo}%

\theoremstyle{thmstylethree}%

\begin{document}

\title{Exploring Possibilities of AI-Powered Legal Assistance in Bangladesh through Large Language Modeling}

%%=============================================================%%
%% GivenName	-> \fnm{Joergen W.}
%% Particle	-> \spfx{van der} -> surname prefix
%% FamilyName	-> \sur{Ploeg}
%% Suffix	-> \sfx{IV}
%% \author*[1,2]{\fnm{Joergen W.} \spfx{van der} \sur{Ploeg} 
%%  \sfx{IV}}\email{iauthor@gmail.com}
%%=============================================================%%

\author*[1]{\fnm{Azmine Toushik} \sur{Wasi}}\email{azmine32@student.sust.edu}
\equalcont{These authors contributed equally to this work.}
\author[1]{\fnm{Wahid} \sur{Faisal}}\email{wahid56@student.sust.edu}
\equalcont{These authors contributed equally to this work.}
\author[2]{\fnm{Mst Rafia} \sur{Islam}}\email{2030391@iub.edu.bd}
\equalcont{These authors contributed equally to this work.}
\author[3]{\fnm{Mahathir Mohammad} \sur{Bappy}}\email{mmbappy@lsu.edu}

\affil[1]{\orgdiv{Industrial and Production Engineering}, \orgname{Shahjalal University of Science and Technology}, \orgaddress{\street{University Ave}, \city{Sylhet}, \postcode{3114}, \state{Sylhet}, \country{Bangladesh}}}
\affil[2]{\orgdiv{Law}, \orgname{Independent University}, \orgaddress{\street{Bashundra R/A}, \city{Dhaka}, \postcode{1229}, \state{Dhaka}, \country{Bangladesh}}}
\affil[3]{\orgdiv{Mechanical and Industrial Engineering}, \orgname{Louisiana State University}, \orgaddress{\street{1146 Pleasant Hall}, \city{Baton Rouge}, \postcode{70803}, \state{LA}, \country{USA}}}
%%==================================%%
%% Sample for unstructured abstract %%
%%==================================%%

\abstract{
\textbf{Purpose:} Bangladesh's legal system struggles with major challenges like delays, complexity, high costs, and millions of unresolved cases, which deter many from pursuing legal action due to lack of knowledge or financial constraints. This research seeks to develop a specialized Large Language Model (LLM) to assist in the Bangladeshi legal system.

\textbf{Methods:} We created \texttt{UKIL-DB-EN}, an English corpus of Bangladeshi legal documents, by collecting and scraping data on various legal acts. We fine-tuned the \texttt{GPT-2} model on this dataset to develop \texttt{GPT2-UKIL-EN}, an LLM focused on providing legal assistance in English.

\textbf{Results:} The model was rigorously evaluated using semantic assessments, including case studies supported by expert opinions. The evaluation provided promising results, demonstrating the potential for the model to assist in legal matters within Bangladesh.

\textbf{Conclusion:} Our work represents the first structured effort toward building an AI-based legal assistant for Bangladesh. While the results are encouraging, further refinements are necessary to improve the model's accuracy, credibility, and safety. This is a significant step toward creating a legal AI capable of serving the needs of a population of 180 million.
}

%Bangladesh's legal system faces major challenges, including delays, complexity, high costs, and millions of pending cases, leading many to avoid legal action due to ignorance or financial constraints.  Motivated by recent progress in Large Language Model-based assistants, we aim to develop a specialized LLM for Bangladeshi legal assistance. We introduce \texttt{UKIL-DB-EN}, an English legal corpus dataset for Bangladesh, and \texttt{GPT2-UKIL-EN}, an LLM designed for Bangladeshi legal assistance in English.We collected and scraped data on legal acts to create \texttt{UKIL-DB-EN}, and fine-tuned the \textit{GPT-2} model on this dataset to develop \texttt{GPT2-UKIL-EN}, designed for legal assistance in English. We rigorously evaluated our model using semantic assessments, including case studies with expert opinions. Our primary findings mark a crucial first step in developing a reliable legal AI for a country of 180 million people, though further work is needed to enhance the model's accuracy and credibility.


%%\pacs[JEL Classification]{D8, H51}

%%\pacs[MSC Classification]{35A01, 65L10, 65L12, 65L20, 65L70}

\maketitle

% AI & SOCIETY  https://link.springer.com/journal/146
% Artificial Intelligence and Law https://link.springer.com/journal/10506
%  Ethics and Information Technology https://link.springer.com/journal/10676

\section{Introduction}
%Legal Natural Language Processing (NLP) is transforming the legal industry by automating the extraction and analysis of information from complex documents like statutes and contracts. Recent advancements have improved tools for document review and legal research, though challenges in legal reasoning persist. The field is growing with increased publications and sophistication, and ongoing developments aim to address current limitations and broaden its applications \cite{Frankenreiter_Nyarko_2023,zhong-etal-2020-nlp}. However, A major challenge for Legal NLP is the variability in legal systems across countries and states. Penal codes and legal norms differ widely, reflecting regional and communal values. AI systems must adapt to these differences to ensure accuracy and relevance, as overlooking regional information can lead to misapplications of the law \cite{katz2023naturallanguageprocessinglegal}.

Legal Natural Language Processing (NLP) is transforming the legal industry by automating the extraction and analysis of information from complex documents such as statutes, contracts, case law, and legal opinions. These advancements are revolutionizing tasks like document review, contract analysis, and legal research, significantly reducing the time and cost associated with legal procedures \cite{lai2023largelanguagemodelslaw,Yan2023}. Tools that leverage NLP can swiftly parse through vast amounts of legal text, identifying key clauses, obligations, and relevant case precedents, providing a much-needed solution in areas like contract drafting, litigation prediction, and regulatory compliance \cite{Frankenreiter_Nyarko_2023,zhong-etal-2020-nlp}. Recent advancements in machine learning, particularly transformer-based models like BERT and GPT, have further enhanced the ability of NLP systems to understand legal texts in context, producing more accurate and relevant insights for legal practitioners \cite{martin2024bettergptcomparinglarge,jiang-etal-2024-leveraging}. 

Despite these advancements, challenges in legal reasoning and the application of AI to complex legal scenarios persist \cite{Yan2023,EnasMohamedAliQuteishat2024}. Legal reasoning often requires a deep understanding of context, precedent, and the intent behind legal provisions, which current NLP models struggle to grasp fully \cite{Zhang2020}. Moreover, the field is growing rapidly, with an increasing number of publications exploring sophisticated methods to tackle these challenges, such as neural-symbolic approaches and domain-specific knowledge graphs. Ongoing developments in Legal NLP aim not only to improve these systems' understanding of legal texts but also to broaden their applications to more complex legal reasoning and cross-jurisdictional cases, overcoming some of the current limitations \cite{Yan2023,ChidiogoUzoamakaAkpuokwe2024,Zhang2020}. A major challenge for Legal NLP, however, is the variability in legal systems across countries and states. Penal codes, legal doctrines, and norms vary widely, reflecting regional, cultural, and communal values. Legal language itself is often hard to understand and subject to interpretation, requiring models to adapt to different jurisdictions to ensure both accuracy and relevance \cite{ChidiogoUzoamakaAkpuokwe2024}. Overlooking these regional distinctions can lead to the misapplication of law, erroneous advice, or incorrect case predictions, posing a significant risk to legal outcomes. Thus, the development of NLP systems for legal applications must account for the intricate and context-dependent nature of legal systems worldwide, ensuring that models are both adaptable and interpretable across varied legal environments \cite{katz2023naturallanguageprocessinglegal,EnasMohamedAliQuteishat2024,Zhang2020}.

\begin{figure}[t] 
\centering {\includegraphics[scale=0.3]{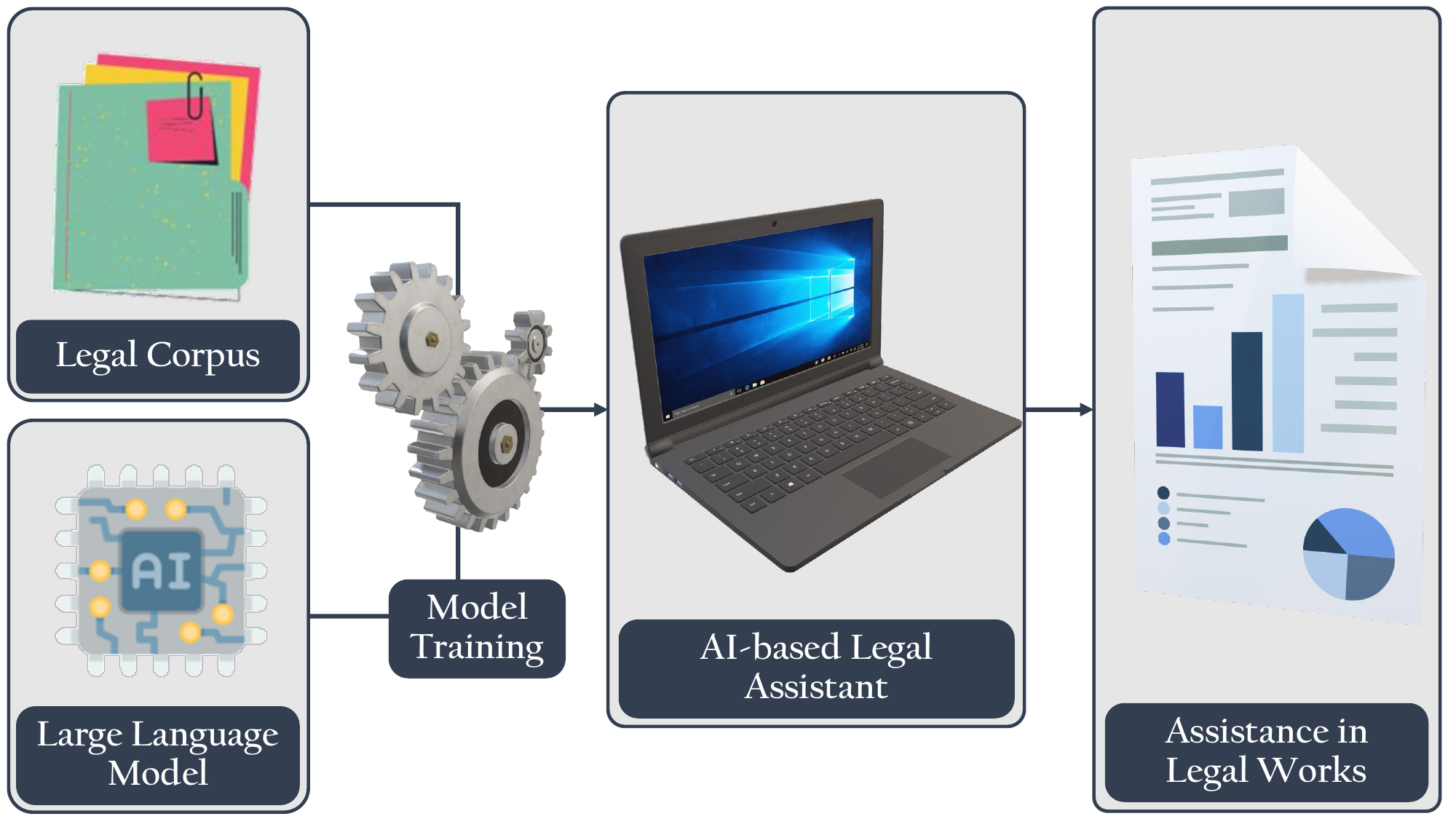}}
\caption{Our approach for exploring AI-powered legal assistance in Bangladesh through LLMs.}
\label{fig:WHAT}
\end{figure}

%The legal system in Bangladesh faces severe challenges, including extensive delays, complex procedures, and high costs that deter many from seeking justice. With a backlog of over 3.7 million cases in 2021, frustration and distrust are prevalent \cite{cegagw43gar3}. Many individuals avoid legal recourse due to ignorance or complexity, while others cannot afford representation, leaving them vulnerable. A Large Language Model (LLM)-based system could significantly improve this situation by automating legal assistance, simplifying legal jargon, and providing accessible, affordable support. By streamlining document preparation, legal research, and case management, such a system could reduce delays and costs, enhance public understanding of legal processes, and broaden access to justice for marginalized populations \cite{sgrsrg32524ef34}.
Legal system in Bangladesh faces severe and longstanding challenges, including significant delays, complex procedural requirements, police harassment, inadequacies in legal provisions, and prohibitive legal costs \cite{20egfesaw24,Raj202wfdf4,Islam2024afeasg,AKTER2017affaf}, which collectively deter many individuals from seeking justice. With a backlog exceeding 3.7 million cases as of 2021 \cite{cegagw43gar3}, the judicial system is overburdened, leading to frustration and widespread distrust in the effectiveness of legal recourse. Cases often take years, if not decades, to reach resolution, leaving many plaintiffs without timely justice \cite{20egfesaw24}. The sheer complexity of legal procedures further complicates access to justice, as individuals unfamiliar with legal jargon or processes find themselves unable to navigate the system without professional legal representation \cite{20egfesaw24,Islam2024afeasg}. Unfortunately, the high cost of legal services means that many in Bangladesh, especially those from lower-income or marginalized communities, cannot afford adequate representation, leaving them vulnerable to exploitation and injustice \cite{Islam2024afeasg}. As a result, many citizens avoid engaging with the legal system altogether, exacerbating inequality and undermining the rule of law \cite{20egfesaw24}. 

We believe, a Large Language Model (LLM)-based system could provide a transformative solution to these issues by automating key aspects of legal assistance, simplifying the complexity of legal language, and offering more accessible, affordable support to the general population. Such a system could assist with tasks like document preparation, case research, and legal consultation, significantly reducing delays and legal fees. By providing citizens with clear, understandable explanations of legal rights and procedures, an LLM-based assistant could empower individuals to engage with the legal system more effectively, ultimately broadening access to justice. Additionally, an AI-based legal assistant could streamline administrative processes within the legal system itself, reducing case backlogs and improving case management efficiency. This could help alleviate some of the systemic delays, ensuring that justice is not only available but also delivered in a timely manner. Such advancements could be particularly beneficial for marginalized populations, who are often excluded from formal legal processes due to financial or educational barriers. By democratizing access to legal information and services, AI-driven tools have the potential to create a more equitable legal system, promoting social justice and fairness in Bangladesh’s legal landscape \cite{sgrsrg32524ef34}.

Motivated by these challenges, we explore the development of a specialized LLM for Bangladeshi legal assistance. To design this model, we first collected data on various legal acts and sections from publicly available open-access government sources and scraped additional data to create \texttt{UKIL-DB-EN} (English meaning of \textit{'Ukil'} is \textit{'Lawyer'}), a comprehensive corpus of Bangladeshi legal documents. We then fine-tuned the \texttt{GPT-2} \cite{radford2019languageGPT2} model on the \texttt{UKIL-DB-EN} dataset in a question-and-answer format to develop \texttt{GPT2-UKIL-EN}, an LLM designed for Bangladeshi legal assistance in English. Our work has been evaluated through various quality and quantitative assessments, including case studies and expert comparisons, to ensure accuracy and reliability. The primary findings show promising results, but further work is needed to improve the model's accuracy, credibility, and safety. Our work is the first ever structured study ever done on Bangladesh context to develop a AI based legal assistant. We believe this work serves as an important first step toward developing a capable legal AI that can assist a developing country with a population of 180 million.

Our core contributions in this work are summarized as below:
\begin{itemize}
    \item We developed \texttt{UKIL-DB-EN}, a comprehensive corpus of Bangladeshi legal documents, by collecting data from open-access government sources and scraping additional legal information.
    \item We fine-tuned the \texttt{GPT-2} model on the \texttt{UKIL-DB-EN} dataset to create \texttt{GPT2-UKIL-EN}, a specialized LLM designed for Bangladeshi legal assistance in English.
    \item Our model was evaluated using both quantitative assessments and case studies, including expert comparisons, to ensure accuracy and reliability.
    \item This is the first structured study focused on developing an AI-based legal assistant specifically for the Bangladeshi legal context, aiming to assist a population of 180 million.
\end{itemize}

This research was conducted in a low-resource setting with limited computational power, preventing us from working with very large models. While Bangla (Bengali) is the primary language in Bangladesh, the legal system uses both Bangla and English—English predominantly in federal high courts or supreme courts, and both languages in lower courts. Due to the significant computational requirements of multilingual LLMs, they were beyond the scope of this study. We focused on training our model solely on English data using an English-based model as part of an exploratory study. This approach, despite limitations, has shown promising potential. RAG-based systems are also out of scope of this work due to lack of computational capabilities. Training larger multilingual models is out of scope for this work, but we believe future studies can build on the foundation we have laid.

In the following sections, we will cover the following topics: related works in Section \ref{sec:RelatedWorks}, dataset development processes in Section \ref{sec:dataset-dev}, model design and development processes in Section \ref{sec:model-dev}, and model performance analysis in Section \ref{sec:RelatedWorks}. We also have developed some cases for legal case studies with the model and evaluated them with expert opinions, as described in Section \ref{sec:case-studies}. We also discuss potential societal impact of the work, limitations of the study and potential future works in Section \ref{sec:discussions}.

\section{Related Works} \label{sec:RelatedWorks}
Recent advancements in Legal LLMs have significantly improved legal tasks such as document analysis \cite{martin2024bettergptcomparinglarge,ygeayhge574yhw5yq4ayq3}, legal text comprehension, and case retrieval \cite{qin2024exploringnexuslargelanguage,Homoki2024wfrwef}, offering faster and more cost-effective solutions than human practitioners \cite{sun2024lawluochineselawfirm}. 
 Some studies integrate LLM-based methods with specialized prompts and curated datasets for LegalAI tasks \cite{huang2024optimizingnumericalestimationoperational}, while others evaluate architectures like Mistral and Gemma for judicial entity extraction and legal retrieval tasks \cite{hussain2024largelanguagemodelsjudicial}. A few works delve into Retrieval-Augmented Generation (RAG) and agent methodologies to improve governance applications and long-form legal question answering \cite{mamalis_kalampokis_fitsilis_theodorakopoulos_tarabanis_2024}. Additionally, several papers investigate the use of LLMs for complex legal concepts through storytelling \cite{jiang-etal-2024-leveraging} and the implementation of prompt templates for structuring legal text processing \cite{deKinderen2024}. The limitations of LLMs, such as their stochastic nature and lack of fact-checking capabilities \cite{sg4ayhgyha45y}, are also discussed, highlighting the need for justifiable and evidence-supported AI models \cite{Louis2024}. Finally, some works emphasize the importance of tailoring LLMs to small law firms and specific legal contexts to optimize their utility and operational efficiency \cite{Homoki2024}.
The introduction of multilingual legal datasets, like MultiLegalPile \cite{qin2024exploringnexuslargelanguage}, expands LLMs' capabilities across various legal systems and languages, promoting inclusivity. However, challenges remain, including biases \cite{wasi2024exploringbengalireligiousdialect}, creativity \cite{wtgwatg4tt4a}, hallucination issues \cite{Dahl20segfeg24}, content ownership \cite{rw3wgt4eteetget4} and the need for customization to handle legal complexities. Ongoing research focuses on improving interpretability, ethical use, and specialized training, aiming to responsibly integrate LLMs into legal workflows, enhancing access to justice and legal services.

Our work focuses on developing a specialized LLM tailored to Bangladesh's legal system, addressing local challenges such as access to justice and high costs. It also includes creating a jurisdiction-specific dataset and evaluating the model through practical case studies, setting it apart from broader advancements in related research.

\section{Developing \texttt{UKIL-DB-EN} Corpus} \label{sec:dataset-dev}
In this section, we outline the data collection and curation methodology for \texttt{UKIL-DB-EN}, along with its corpus structure and descriptive statistics. 

\subsection{Data Collection and Corpus Development}
In this section, we outline the comprehensive data collection and curation process undertaken for the development of the \texttt{UKIL-DB-EN} dataset. This process involved systematic scraping of legal texts from an open-access government portal and meticulous preprocessing to ensure the quality and relevance of the data for our legal language model.
\subsubsection{Data Collection}
To create the \texttt{UKIL-DB-EN} dataset, we initiate a systematic data collection process by scraping an open-access government portal\footnote{http://bdlaws.minlaw.gov.bd/} dedicated to providing legal information in Bangladesh. This portal serves as a comprehensive repository of legal texts, including statutes, regulations, and official legal documents that are essential for understanding the legal landscape of the country. 

Using a Python-based web-scraping tool, we develop a customized script to navigate through the various sections of the website efficiently. The tool was designed to target specific URLs that contained relevant legal documents, allowing us to extract a diverse array of legal texts. Our scraping process involved identifying and collecting information from multiple categories, such as civil laws, criminal laws, administrative regulations, and various legal acts, ensuring a broad representation of Bangladesh's legal framework.

\subsubsection{Data Curation}
Following the extraction of raw data, we engaged in a comprehensive preprocessing phase aimed at enhancing the quality and relevance of the dataset for model training. This included several key steps:
\noindent
\textbf{Data Cleaning.} We meticulously removed repealed acts and sections to ensure that only current and applicable legal texts were included in the dataset. This step was critical to maintain the integrity and reliability of the information, as outdated laws could mislead users and negatively impact the model's performance.

\noindent
\textbf{Noise Reduction.} During the extraction process, we encountered various unwanted characters and formatting inconsistencies. We implemented string manipulation techniques to eliminate noise, including special characters, irrelevant whitespace, and other extraneous elements that could interfere with natural language processing tasks.

\noindent
\textbf{Standardization.} We standardized the formatting of the legal texts to ensure uniformity across the dataset. This included converting all text to a consistent case, structuring citation formats, and ensuring that the presentation of legal clauses and provisions adhered to a clear and recognizable structure.

\noindent
\textbf{Verification and Validation.} To ensure the accuracy and reliability of the collected information, all data was rigorously checked and verified by the authors. We cross-referenced the scraped texts with official sources and documentation to confirm their validity. This validation process was crucial in establishing the credibility of the dataset, as it forms the foundation for subsequent model training and application.

Through these meticulous data collection and curation efforts, we have established the \texttt{UKIL-DB-EN} dataset as a valuable resource for developing a specialized legal language model tailored to the unique context of Bangladeshi law. This dataset is not only a reflection of the current legal framework but also a vital step towards improving access to legal information and assistance for the people of Bangladesh.

\subsection{Data Structure} \label{sec:apx:more-data-info}
We have organized the related acts and sections hierarchically within the \textit{JSON} structure for efficient information retrieval. At the top level, the key \textit{act} contains essential information about the act itself, including a unique \textit{id},  \textit{name} of the act, a boolean \textit{repelled} indicating whether the act has been repealed, the \textit{text} containing details of the act, the \textit{published\_date}, a list of \textit{related\_act} IDs that connect the act to other \textit{acts}, \textit{lower\_text} array for additional notes, the \textit{num\_of\_sections} indicating the total sections present, and an array of \textit{sections}. Each section within the \textit{sections} array is represented as an object with keys such as \textit{section\_id}, \textit{name}, and \textit{details} that provide specific information about that section. Additionally, each section may include a \textit{related\_acts} array for further connections to other acts and an \textit{act\_id} linking it back to the main act. Data sample is available in Appendix, in Table \ref{tab:sample}.

\subsection{Descriptive Statistics}
We have collected information on a total of 595 Acts, which include approximately 18,023 sections. On average, each act contains about 24 sections. The mean character length for act names is 50.30, while act details have a mean length of 438.37 characters. Section names are shorter, averaging 38.07 characters, reflecting their concise titles. In contrast, section details are more comprehensive, with an average length of 736.69 characters, indicating detailed descriptions and provisions.

\section{LLM Model : \texttt{GPT2-UKIL-EN}} \label{sec:model-dev}
In this section, we discuss the model fine-tuning process, including prompt design and implementation details such as hyperparameters and other technical specifications.

\subsection{Model Development}
Our work uses pre-trained GPT-2\footnote{https://huggingface.co/openai-community/gpt2-medium} model from HuggingFace, originally developed by \citet{radford2019languageGPT2}. We also use the original tokenizer \texttt{(GPT2Tokenizer} \textit{class}) to tokenize our data, from the \texttt{transformers} library \cite{wolf2020huggingfaces}. 
To train the model with legal information, we have developed prompts using a simple instruction-tuning approach. For acts, we have used the prompt: \textit{What do you know about "\#act\_name", "year", Bangladesh?} The model has responded with a concatenation of all section details in text format. For sections, we have used the prompt: \textit{What do you know about "\#section\_name" from "\#act\_name", "year", Bangladesh?} The model has provided the specific section detail texts. Upon acceptance, we will publicly release both the text corpus and the trained model.

\subsection{Implementation Details}
All the random seeds used are 42. We configured the model with a batch size of $64$ and further divided it into micro-batches of size $8$ for computation, leading to a gradient accumulation step of $8$. We have trained the model for 13 epochs with a learning rate of \(3 \times 10^{-4}\) with warm-up steps of 2 and set $fp16$ to for computational efficiency. The validation set comprised 2000 samples from 18,488 total prompts. 
We have implemented Low-Rank Adaptation (LoRA) for efficient training, with rank $3$, alpha set $16$, and task type \texttt{CAUSAL\_LM}. Targeting modules used are \textit{attn.c\_attn, mlp.c\_fc}, with a dropout rate of 0.1 and bias set to \texttt{none}. For quantization, we have employed \texttt{nf4} \textit{double quantization}, with compute type \texttt{float16}. For the tokenizer parameters, we have used the maximum length of $768$, set padding strategy set to \textit{max\_length} and truncation to true.

\begin{table}[t] 
\centering
\caption{Experimental results.}\label{tab:QuanEval}
\begin{tabular}{lcccc}
\toprule
Model & Parameters & Fine-tuned? & Cosine Sim. ($\uparrow$) & Jaccard Sim. ($\uparrow$) \\ 
\midrule
Mistral-7b  & 7B & No & 0.446 & 0.122  \\
Gemma-2b  & 2B & No & 0.436   & 0.113  \\ 
GPT-2 Medium  & 0.345B & No  & 0.178 & 0.062  \\ 
GPT2-UKIL-EN (Ours)  & 0.345B & Yes & \textbf{0.515} & \textbf{0.133}  \\
\bottomrule
\end{tabular}
\end{table}

\section{Model Performance Analysis} \label{sec:model-performance-analysis}
In this section, we assess the performance of \texttt{GPT2-UKIL-EN} through a combination of semantic analysis and error analysis. This evaluation aims to identify the model's strengths and weaknesses, providing insights into its effectiveness in delivering accurate legal assistance.

\subsection{Semantic Similarity Analysis}
To assess the model semantic similarity quantitatively, we use Cosine similarity and Jaccard index to measure the text similarity between the original text and the output of different models, as presented in Table \ref{tab:QuanEval}. These metrics help evaluate how well the models' generated outputs match the original texts, which is important in legal work.
\texttt{GPT2-UKIL-EN}, our fine-tuned model, significantly outperforms the others, achieving the highest scores in both Cosine similarity (0.515) and Jaccard index (0.133). This indicates that fine-tuning on the \texttt{UKIL-DB-EN} dataset improves model performance in capturing legal text semantics and overlap. While \texttt{Mistral-7b} and \texttt{Gemma-2b} show competitive results without fine-tuning, their scores fall short, suggesting the value of domain-specific fine-tuning for legal tasks. \texttt{GPT-2 Medium} performs the lowest, emphasizing the impact of both model size and fine-tuning. 

Additionally, we observe a general correlation between model size and text similarity scores: as model size increases, the similarity scores tend to improve. Despite being 6-10x smaller than the larger models, our fine-tuned model performs remarkably well, showing a significant improvement over the original \texttt{GPT-2} model. This demonstrates that our fine-tuning process substantially enhanced the model's capabilities. Further insights and concerns on this are discussed in \textit{Limitations}.

\subsection{Error Analysis}
The error analysis presented in Table \ref{tab:ErrorAnalysis} reveals significant discrepancies between the expected answer and our model's response regarding the "Power to make rules" section from The Pensions Act, 1871 of Bangladesh. The expected answer outlines specific powers granted to the Parishad for rule-making related to the Academy’s membership, functions, and terms of service, detailing practical aspects and administrative functions. In contrast, our model's response inaccurately describes rule-making authority related to a different chapter, emphasizing rules that were supposed to be in force from March 1979 and including information about the Board and Education Act amendments. This response introduces unrelated content such as rule validity and Board establishment, diverging from the actual provisions of the 1871 Act. The model's failure to accurately identify and reproduce the specific administrative details reflects a misunderstanding of the context and subject matter, indicating a need for improved contextual comprehension and accurate content retrieval. This highlights the model’s struggle with aligning its responses to the specific legal and historical context required, suggesting areas for refinement in handling detailed legislative texts.

\begin{table}
\centering
\caption{Error Analysis.}\label{tab:ErrorAnalysis}
\begin{tabular}{p{2cm}p{4.5cm}p{5.5cm}}
\toprule
Question & Expected Answer & Our Model  \\ 
\midrule
What do you know about "Power to make rules" section from The Pensions Act, 1871 Bangladesh?
& The Parishad may, with prior approval of the Government, make rules for carrying out the purposes of this Ordinance. (2) Without prejudice to the generality of the power conferred by sub-section (1), such rules may provide for all or any of the following matters, namely:- (a) Membership of the Academy; (b) functions and conduct of work of different Divisions of the Academy; ... (to be continued)
& The Government may, by notification in the official Gazette, make such rules as may be necessary for carrying into effect the provisions of this Chapter. (2) The rules referred to in sub-section (1) shall come into force on the first day of March, 1979, and shall remain in force for a period of one year from the date on which they are promulgated. Explanation.- (a) In this section, unless there is anything repugnant in any of the subject matter contained therein,- (i) "Board" means the Board constituted under section 3; and ... (to be continued)
 \\
\bottomrule
\end{tabular}
\end{table}

\section{Case Study and Comparisons} \label{sec:case-studies}
To better understand and assess the model, we developed three cases and collected expert opinions on these cases, on different statements and comments from the experts.

\subsection{Case Design}
We developed three distinct legal cases for expert evaluation: a property dispute (Case 1, Hard), an illegal possession case under special powers act (Case 2, Easy), and a murder charge (Case 3, Medium), which from different areas and difficulty. In the property  case, the model was expected to verify the authenticity of ownership documents for a 500-square-foot plot in Dhaka. In the illegal possession case, involving Imran Hossain under the Special Powers Act, 1974, the model was tasked with determining whether his actions posed a significant public safety risk, justifying prosecution. Lastly, in the murder case, the model was expected to analyze the evidence against Amirul Islam, charged under Section 302 of the Penal Code, and decide if the crime fell under any exceptions to the murder charge. 
Table \ref{tab:ApxCaseStudies} shows three different cases developed by us, and their responses solved by the model. The expert opinions are shortly presented at Table \ref{tab:CaseStudiesResults}, detailed collected results are available in Table \ref{tab:CaseStudiesResults-CS}. 
In Table \ref{tab:CaseStudiesResults-CS}, each asterisk (*) denotes one expert's opinion on that particular case and statement.

\begin{sidewaystable}
\centering
\caption{Case Design and Development (All Cases: Question and Answer).}\label{tab:ApxCaseStudies}
    \begin{tabular*}{\textheight}{@{\extracolsep\fill}p{1cm}p{8cm}p{9cm}}
\toprule
No & Case & Answer by \texttt{GPT2-UKIL-EN} \\ 
\midrule
1
& Mr. Ahmed Rahman and Mr. Tariq Karim are involved in a property dispute over a 500-square-foot plot in Dhaka. Rahman claims ownership based on a sale deed from 2015, with the land developed and maintained by him. Karim argues that the land, which his family originally owned, was fraudulently sold to Rahman. Karim has presented a copy of the original land deed and accuses the sale transaction of being invalid. The dispute centers on the validity of the documents and the legitimacy of the ownership claims.
How the Court can find the valid documents?
& Answer:   The Court may, if it thinks fit, find that there is no valid document for the purposes of this section. (2) Where a document is found to be invalid under sub-section (1), it shall be deemed to have been so invalid that it cannot be used as evidence in any civil or criminal case, and shall not be admissible in evidence at any trial or hearing in respect of any matter arising under this Act or any other law for which it is not an offence under the Code of Criminal Procedure, 1898, or under any rule made under that Code for a period of one year from the date of its publication in the Officia...\\ \hline
2  
& Imran Hossain, a small-scale trader, is charged under the Special Powers Act, 1974, for illegal possession and sale of controlled chemicals. The prosecution claims that Imran was found with a large quantity of restricted chemicals, which he was selling without proper authorization. This violation addresses the actions prejudicial to the security of the state, as the unregulated sale of these chemicals could pose risks to public safety and national security. 
Do Imran’s actions warrant prosecution under the Special Powers Act due to their potential threat to public welfare?
& Answer:  In order to protect the public health and safety, the Government may, by notification in the official Gazette, provide for such punishment as it deems fit. (2) Notwithstanding anything contained in any other law for the time being in force, an offence under sub-section (1) shall be punishable with imprisonment of either description for a term which may extend to three years, and shall also be liable to fine, if any, not exceeding one hundred thousand Taka, or with fine of not less than two hundred taka and not more than five hundred kuna. Explanation:- (a) In this section, unless there is anything repugnant in its context,- (i) "controlled chemical" means any substance which is capable of being used... (out of length)  \\ \hline
3 
& Amirul Islam is charged with murder under Section 302 of the Penal Code of Bangladesh. On July 15, 2024, he allegedly stabbed Zubayer Ali during an argument over a business deal. Witnesses saw Islam at the scene, and forensic evidence links him to the crime. Islam confessed to the murder but the defense claims his confession was forced. 
Now the question is, did Amirul Islam committed Murder?
& Answer:  If the answer is in the affirmative, the accused shall be punished with imprisonment of either description for a term not exceeding three years, or with fine which may extend to one thousand Taka. (2) If, after having been convicted under sub-section (1), the Court is satisfied that there is no reasonable ground for doubt, it may, if it thinks fit, direct that the case be referred to a Special Court for the purpose of determining the punishment to be inflicted. The Court shall make such order as may be necessary for carrying out the purposes of this section.  (3) Notwithstanding anything contained in this Chapter or any other law for which it is not repugnant, a person convicted of Murder under section 302 shall not be... (out of length)
  \\
\bottomrule
\end{tabular*}
\end{sidewaystable}

\begin{table}[t] 
\centering
\caption{Case Study Results.}\label{tab:CaseStudiesResults}
\begin{tabular}{p{8cm}cccc}
\toprule
Statement & C1 & C2 & C3 & Average\\ 
\midrule
1. I think the response provided by the tool is accurate. 
& 4.25 & 5.25 & 2.25 & 3.92 \\
2. I think the answer given by the tool is clear and precise. 
& 5.00 & 5.25 & 3.00 & 4.42\\
3. The tool reasons well with the given context. 
& 5.00 & 6.00 & 5.00 & 5.33\\
4. I believe the tool's approach and reasoning are commendable. 
& 5.50 & 5.50 & 5.25 & 5.42\\
5. I feel that the response addressed the question effectively. 
& 5.50 & 4.75 & 4.00 & 4.67\\
6. The tool accurately processes relevant acts and sections. 
& 4.50 & 4.75 & 3.75 & 4.33\\
7. The approach and style of the tool's output are good. 
& 5.50 & 5.75 & 5.50 & 5.58\\ \hline
Per case average
& 5.00 & 5.32 & 4.11 & 4.81 \\
\bottomrule
\end{tabular}
\end{table}

\subsection{Study Design}
We conducted a case study with five legal experts, consisting of four male and one female law faculty members\footnote{Identities will be revealed in the final version, E4 didn't fill up the survey form, but shared a lot of insights}, to evaluate our model’s performance. We presented our motivation, explained the model's components and functionality, and demonstrated three cases (Table \ref{tab:ApxCaseStudies}) solved by the model. The experts were asked to assess the model based on seven affirmative statements (questions are available in Table \ref{tab:CaseStudiesResults} and \ref{tab:CaseStudiesResults-CS}) designed to evaluate aspects such as accuracy, quality, reasoning ability, approach, and writing style. Each statement was rated on a scale from 1 to 7: Strongly Disagree (1), Disagree (2), Somewhat Disagree (3), Neutral (4), Somewhat Agree (5), Agree (6), and Strongly Agree (7). All the statements are affirmative, so more than 4 means positive and less than 4 means negative. Along with the study, the experts provided valuable insights verbally and in written.

\noindent
\textbf{Information on Experts.}
The experts are law faculty members, aged between 32 and 60, including three lecturers, one Assistant Professor, and one Professor. Their diverse academic ranks and experience offer a broad range of insights and feedback on the model's performance. Notably, E4 did not provide opinions on the cases but shared extensive insights on the work's potential. As a result, there are four responses for each case and statement. We will reveal the experts' identities in the final version. The experts are neutral and have no vested interest in the work.

\subsection{Performance Analysis}
\noindent
\textbf{Overall Performance Analysis.}
From the expert opinions presented in Table \ref{tab:CaseStudiesResults}, our model \texttt{GPT2-UKIL-EN} excels in reasoning and approach, with high scores indicating strong logical thinking and effective methods. However, it struggles with accuracy and clarity, particularly in complex scenarios, as reflected in lower scores for these aspects. While the model performs well in simpler cases, its accuracy and clarity suffer in more intricate contexts, revealing a need for improvement. The highest score for approach and style shows that the model generally presents information effectively, but its inconsistent performance in accuracy highlights areas for improvement in diverse tasks.

\noindent
\textbf{Case-wise Analysis.}
\texttt{GPT2-UKIL-EN}'s lower accuracy score (4.25) in Case 1 highlights difficulties with complex property document verification, likely due to the challenge of validating multiple layers of legal documentation. Clarity and precision also suffered (5.00), indicating issues with ambiguous contexts. Despite reasonable ratings for reasoning (5.00) and approach (5.50), the model struggled with the complexity, impacting overall effectiveness.
In Case 2, the model achieved its highest scores for accuracy (5.25) and clarity (5.25), reflecting effective handling of straightforward public safety risk assessments. The simplicity of the case allowed for clear, precise responses. The highest reasoning score (6.00) and commendable approach (5.75) underscore the model's strong performance in simpler scenarios.
For Case 3, the model's lowest accuracy score (2.25) indicates difficulties with analyzing complex evidence and exceptions to a murder charge. The lower clarity score (3.00) further highlights challenges with intricate legal details. Despite good reasoning (5.00) and approach (5.25) scores, the model's effectiveness was diminished due to the case's complexity, revealing need of improvement in handling complex scenarios.

\noindent
\textbf{Qualitative Analysis on Expert Opinions.}
The experts raised several key concerns and insights about the model. Issues identified include an information gap, noted by experts E2, E3, and E4, suggesting the need for more comprehensive data or details. Experts E1, E2, and E3 recognized the value of the model's core idea but also pointed out that the language used should be simplified for better understanding, as highlighted by E1, E2, and E5. Additionally, E1, E2, and E4 emphasized the need to make responses more context-specific and to simplify answers overall. Addressing these concerns involves enhancing the model’s contextual relevance, simplifying the language and responses, and bridging the information gap to improve clarity and effectiveness. All experts praised the overall idea, study, and effort, and they have promised to contribute to future surveys and analysis.

\section{Discussion} \label{sec:discussions}
We believe our \texttt{GPT2-UKIL-EN} model represents a valuable step forward in Legal NLP for Bangladesh, offering a tailored approach that could help reduce costs and accelerate legal procedures. This model not only aims to streamline the legal process but also strives to democratize access to legal information, thereby empowering individuals who have historically faced barriers in navigating the legal system. By providing clear and understandable explanations of legal terms and processes, our model can significantly enhance public understanding, making the legal system more approachable for those who may feel intimidated or overwhelmed by its complexity.

While our approach has the potential to decrease the case backlog and improve access to legal assistance, particularly for marginalized communities, we acknowledge that the model still has several limitations. Issues such as incomplete legal coverage, potential inaccuracies in generated content, and a reliance on existing datasets could impede the model's effectiveness in certain contexts. However, we believe that with further exploration and increased computational resources, these challenges can be addressed, leading to continuous improvements in the model's accuracy and reliability.
The societal impact of \texttt{GPT2-UKIL-EN} could be transformative. By facilitating easier access to legal information, it empowers individuals to make informed decisions regarding their legal rights and responsibilities. This is particularly crucial in Bangladesh, where many individuals may avoid seeking legal recourse due to a lack of understanding or financial constraints. By simplifying legal jargon and enhancing public understanding of legal processes, our model contributes to a more equitable legal system that can support the rights of all citizens, regardless of their socio-economic status.

Globally, the implications of our work extend beyond the boundaries of Bangladesh. The methodologies and frameworks established through this project can serve as a model for other countries facing similar challenges in their legal systems. As legal systems around the world increasingly adopt AI and NLP technologies, our approach may provide valuable insights into developing multilingual legal assistants tailored to diverse legal contexts. This adaptability underscores the potential for a global impact, as nations can leverage our findings to enhance their own legal frameworks, thereby contributing to the ongoing evolution of legal technology.
Moreover, the use of AI in legal settings can facilitate more efficient court processes, reducing the time and resources spent on legal proceedings. This efficiency could lead to a significant decrease in the overall costs associated with legal disputes, benefiting not only individuals but also the judicial system as a whole. By addressing delays and streamlining case management, our model could help restore public trust in legal institutions, encouraging more individuals to seek justice and ensuring that their rights are upheld.
Overall, we belive, \texttt{GPT2-UKIL-EN} holds promise not just as a tool for legal assistance but as a catalyst for broader societal change. With continued refinement and exploration, this approach could pave the way for similar advancements in other legal contexts facing unique challenges, ultimately fostering a more just and accessible legal landscape worldwide.

\noindent
\textbf{Limitations of the Study and Future Work Possibilities.}
This study was conducted under several constraints, primarily due to limited computational resources. As a result, we were unable to experiment with very large models or state-of-the-art multilingual language models, which are typically required for effective performance in both Bangla and English. Given that the legal system in Bangladesh uses both languages—English predominantly in higher courts and a mixture of Bangla and English in lower courts—this limitation restricts the broader applicability of our model across all legal contexts in Bangladesh.
Furthermore, our exploration was limited to training on English data using an English-based model, which means our model may not fully capture the complexities of the legal processes in lower courts where Bangla is also widely used. Developing and fine-tuning multilingual models that can handle both languages effectively was beyond the scope of this work, but we believe this is a critical area for future research. Additionally, due to resource limitations, we could not explore larger-scale models, which may yield better performance in complex legal reasoning tasks. Despite these limitations, our findings show significant promise and lay the groundwork for future studies to address these challenges.

\section{Conclusion}
The current legal system of Bangladesh faces challenges such as delays, complexity, and high costs, with over 3.7 million cases pending in 2021. LLMs have the potential to ease these issues. Motivated by this, we have developed \texttt{UKIL-DB-EN}, an English legal corpus dataset, and \texttt{GPT2-UKIL-EN}, a large language model for legal assistance. We created \texttt{UKIL-DB-EN} by scraping data on legal acts and fine-tuned the GPT-2 model on this dataset. 
Our work, the first ever effort to create a Bangladesh-focused AI-based legal assistant model, evaluated through various assessments and expert opinions, shows promising results and marks an important first step toward developing a reliable legal AI for a country of 180 million people.

\backmatter

\bmhead{Acknowledgements}
We would like to express our sincere gratitude to all the esteemed members of the expert panel for their valuable time, guidance, and thoughtful feedback throughout the assessment process. Our deepest appreciation goes to Prof. Dr. Muhammad Mahbubur Rahman, Prof. Dr. Liton Chandra Biswas, Mr. Md Al Ifran Hossain Mollah, Mr. Washik Muhammod Istiaz Ezaz, and Ms. Afroza Bilkis for their valuable insights and our model. We are also grateful to Computational Intelligence and Operations Laboratory (CIOL) for all kinds of support and guidance in the work.

\section*{Declarations}

\begin{itemize}
\item Funding : No funding was received to assist with the preparation of this manuscript.
\item Conflict of interest: All authors certify that they have no affiliations with or involvement in any organization or entity with any financial interest or non-financial interest in the subject matter or materials discussed in this manuscript.
\item Ethical Concerns: As we are developing legal AI, the correctness of the model is crucial. Given that our model is in its early stages, there may be issues related to its accuracy and reliability. This raises important ethical concerns, particularly regarding the potential impact of inaccuracies in legal contexts. 
\item Ethics approval and consent to participate: We adhered to all ethical guidelines outlined in the Springer Nature guidelines during data collection, curation, and modeling experiments. We also reported all hyperparameters used in our experiments to ensure reproducibility and transparency, aiming to uphold the highest standards of ethical practice in our research.
Our study also involved working with experts who provided opinions on various legal cases. In accordance with Springer Nature and ethical research guidelines, we ensure that the experts are neutral, with no vested interest in the work, thus maintaining objectivity and impartiality.
\item Consent for publication : All identifying details of participants have been published with informed consent, ensuring compliance with ethical guidelines. Anonymization has been applied where applicable, and no identifying information has been included without proper consent.
\item  Potential Risks and Response: Developing legal AI requires a high level of accuracy due to the critical nature of its applications. As our model is still in its early development phase, there is a risk of inaccuracies and reliability issues that could impact its performance in legal contexts. These potential risks underscore the importance of addressing ethical concerns, as any errors could have significant consequences. To mitigate these risks, we have followed ethical guidelines for data collection, curation, and modeling as specified by ACL. 
%\item Use of Generative AI and AI-assisted Technologies: During the preparation of this work the author(s) used ChatGPT in order to reduce grammatical errors and writing clarity. After using this tool/service, the author(s) reviewed and edited the content as needed and take(s) full responsibility for the content of the published article.
\item Data availability : The \texttt{UKIL-DB-EN} dataset is publicly available in Hugging Face with DOI: \href{https://doi.org/10.57967/hf/3233}{\texttt{10.57967/hf/3233}} \cite{https://doi.org/10.57967/hf/3235} with Apache-2.0 license.
\item Materials availability : The \texttt{GPT2-UKIL-EN} model is publicly available in Hugging Face with DOI: \href{https://doi.org/10.57967/hf/3235}{\texttt{10.57967/hf/3235}} \cite{https://doi.org/10.57967/hf/3233} with Apache-2.0 license.
\item Code availability : All codes are available in GitHub on \href{https://github.com/ciol-researchlab/UKIL}{\texttt{ciol-researchlab/UKIL}} with Apache-2.0 license.
\item Author contribution : A.T.W., W.F., and M.R.I. contributed equally to the project. A.T.W. conceptualized the study, managed data curation, formal analysis, project administration, validation, visualization, made drafts and finalized the manuscript. W.F. led data collection, formal analysis, investigation, methodology, software, resource management, and contributed to writing the manuscript. M.R.I. contributed to data collection, investigation, methodology, case development, interviews, resource management, study validation, and assisted with drafting and reviewing the manuscript. M.M.B. supervised the work, reviewed the manuscript, and provided feedback.
\end{itemize}

\begin{appendices}

\begin{sidewaystable}
    \centering
    \caption{Sample from \texttt{UKIL-DB-EN} Corpus (Sections data)}
    \begin{tabular*}{\textheight}{@{\extracolsep\fill}|p{1cm}|p{2cm}|p{12cm}|p{1cm}|p{1cm}|}
        \toprule
        \textbf{Section ID} & \textbf{Name} & \textbf{Details} & \textbf{Related Acts} & \textbf{Act ID} \\
        \midrule
        1 & Public Accountants to give security & Every public accountant shall give security for the due discharge of the trusts of his office, and for the due account of all moneys which shall come into his possession or control, by reason of his office. &  & 2 \\
        \hline
        2 & Amount and kind of security, and with what sureties & In default of any Act having special reference to the office of any public accountant, the security given shall be of such amount and kind, real or personal, or both, and with such sureties (regard being had to the nature of the office), as shall be required by any rules made or to be made from time to time, by the authority by which each public accountant is appointed to his office. &  & 2 \\
        \hline
        3 & “Public accountant” defined & For the purposes of sections 1 and 2 of this Act, the expression “public accountant” means any person who as Official Assignee or Trustee, or as sarbarahkar, is entrusted with the receipt, custody or control of any moneys or securities for money, or the management of any lands belonging to any other person or persons; and for the purposes of sections 4 and 5 of this Act the expression shall also include any person who, by reason of any office held by him in the service of the State is entrusted with the receipt, custody or control of any moneys or securities for money, or the management of any lands belonging to the Government. &  & 2 \\
        \hline
        5  & Enactments applied to proceedings by and against accountants & All Regulations and Acts now or hereafter to be in force for the recovery of arrears of land-revenue due to Government, and for recovery of damages by any person wrongfully proceeded against for any such arrear shall apply, with such changes in the forms of procedure as are necessary to make them applicable to the case, to the proceedings against and by such public accountant &  & 2\\
        \hline
        13  & Evidence for prosecution and examination of witnesses.Re-examination by prosecutor & The oral and documentary evidence for the prosecution shall then be exhibited; the witnesses shall be examined by or on behalf of the prosecutor and may be cross-examined by or on behalf of the person accused. The prosecutor shall be entitled to re-examine the witnesses on any points on which they have been cross examined, but not on any new matter, without leave of the commissioners, who also may put such question as they think fit &  & 4\\
        \hline
        14178 & Pledge by mercantile agent & Where a mercantile agent is, with the consent of the owner, in possession of goods or the documents of title to goods, any pledge made by him, when acting in the ordinary course of business of a mercantile agent, shall be as valid as if he were expressly authorized by the owner of the goods to make the same; provided that the pawnee acts in good faith and has not at the time of the pledge notice that the pawnor has not authority to pledge. Explanation – In this section, the expressions 'mercantile agent' and 'documents of title' shall have the meanings assigned to them in the Sale of Goods Act, 1930 & 150 & 26\\
       \bottomrule
    \end{tabular*}
    \label{tab:sample}
\end{sidewaystable}

\begin{sidewaystable}
\centering
\caption{Detailed Case Study Results.}\label{tab:CaseStudiesResults-CS}
\begin{tabular*}{\textheight}{@{\extracolsep\fill}|p{.5cm}|p{7cm}|p{1.25cm}|p{1.15cm}|p{1.25cm}|p{1cm}|p{1.15cm}|p{0.75cm}|p{1.25cm}|}
\toprule
Case No & Statement & Strongly Disagree (1) & Disagree (2) & Somewhat Disagree (3) & Neutral (4) & Somewhat Agree (5) & Agree (6) & Strongly Agree (7) \\ 
\midrule
1 & I think the response provided by the tool is accurate.
&  & *  &  & * &* &  *&  \\ \hline
1 & I think the answer given by the tool is clear and precise. 
&  &  & &  & ****&  &  \\ \hline
1 & The tool reasons well with the given context. 
&  &  & *&  & *& ** &  \\ \hline
1 & I believe the tool's approach and reasoning are commendable. 
&  &  & &  & **& ** &  \\ \hline
1 & I feel that the response addressed the question effectively. 
&  &  & & * &* & ** &  \\ \hline
1 & The tool accurately processes relevant acts and sections. 
&  &  &* &*  &* & * &  \\ \hline
1 & The approach and style of the tool's output are good. &  
&  & & * & & *** &  \\ \hline
2 & I think the response provided by the tool is accurate. 
&  &  & &  *& *& ** &  \\ \hline
2 & I think the answer given by the tool is clear and precise. 
&  &  & & * & **& ** &  \\ \hline
2 & The tool reasons well with the given context. 
&  &  & &  & &****  &  \\ \hline
2 & I believe the tool's approach and reasoning are commendable. 
&  &  & &  &** & ** &  \\ \hline
2 & I feel that the response addressed the question effectively. 
& * &  & &  & *& ** &  \\ \hline
2 & The tool accurately processes relevant acts and sections. 
& * &  & &  & *& ** &  \\ \hline
2 & The approach and style of the tool's output are good. 
&  &  & &  & *& *** &  \\ \hline
3 & I think the response provided by the tool is accurate. 
*& * & ** & &  & &  &  \\ \hline
3 & I think the answer given by the tool is clear and precise. 
&  &  & ****&  & &  &  \\ \hline
3 & The tool reasons well with the given context. 
&  &  & * &  &* & ** &  \\ \hline
3 & I believe the tool's approach and reasoning are commendable. 
&  &  & &  & ***& * &  \\ \hline
3 & I feel that the response addressed the question effectively. 
&  &  &** & * & & * &  \\ \hline
3 & The tool accurately processes relevant acts and sections. 
&  & ** & &  & *&*  &  \\ \hline
3 & The approach and style of the tool's output are good.
&  &  & &  & **& ** &  \\ 
\bottomrule
\end{tabular*}
\footnotetext{Note: Each asterisk (*) denotes one expert's opinion on that particular case and statement.}
\end{sidewaystable}

%%=============================================%%
%% For submissions to Nature Portfolio Journals %%
%% please use the heading ``Extended Data''.   %%
%%=============================================%%

%%=============================================================%%
%% Sample for another appendix section			       %%
%%=============================================================%%

%% \section{Example of another appendix section}\label{secA2}%
%% Appendices may be used for helpful, supporting or essential material that would otherwise 
%% clutter, break up or be distracting to the text. Appendices can consist of sections, figures, 
%% tables and equations etc.

\end{appendices}

%%===========================================================================================%%
%% If you are submitting to one of the Nature Portfolio journals, using the eJP submission   %%
%% system, please include the references within the manuscript file itself. You may do this  %%
%% by copying the reference list from your .bbl file, paste it into the main manuscript .tex %%
%% file, and delete the associated \verb+\bibliography+ commands.                            %%
%%===========================================================================================%%

\bibliography{a-our_work}% common bib file
%% if required, the content of .bbl file can be included here once bbl is generated
%%\input sn-article.bbl

\end{document}